\documentclass[runningheads]{llncs}
\usepackage{graphicx}
\usepackage{pifont}
\usepackage{wasysym}
\usepackage{utfsym}
\usepackage{fontawesome}
\usepackage{multirow}
\usepackage{marvosym}

\begin{document}

\title{Semi-Supervised Semantic Segmentation Methods for UW-OCTA Diabetic Retinopathy Grade Assessment\thanks{Supported by Ministry of Higher Education, Malaysia.}}
\titlerunning{Semi-Supervised Segmentation Methods for DR Grade Assessment}
%
\author{Zhuoyi Tan\orcidID{0000-0002-3422-2504} \and
Hizmawati Madzin\textsuperscript{(\Letter)} \orcidID{0000-0003-0098-2086} \and
Zeyu Ding\orcidID{0000-0001-8675-8370} 
}

\authorrunning{Zhuoyi.T et al.}
%
\institute{Faculty of Computer Science and Information Technology, University of Putra Malaysia, Serdang 43400, Malaysia \\
\email{hizmawati@upm.edu.my}\\
}
\maketitle              

\begin{abstract}
People with diabetes are more likely to develop diabetic retinopathy (DR) than healthy people.     However, DR is the leading cause of blindness.    At present, the diagnosis of diabetic retinopathy mainly relies on the experienced clinician to recognize the fine features in color fundus images.    This is a time-consuming task.  Therefore, in this paper, to promote the development of  UW-OCTA DR automatic detection, we propose a novel semi-supervised semantic segmentation method for UW-OCTA DR image grade assessment.  This method, first, uses the MAE algorithm to perform semi-supervised pre-training on the UW-OCTA DR grade assessment dataset to mine the supervised information in the UW-OCTA images, thereby alleviating the need for labeled data.  Secondly, to more fully mine the lesion features of each region in the UW-OCTA image, this paper constructs a cross-algorithm ensemble DR tissue segmentation algorithm by deploying three algorithms with different visual feature processing strategies.  The algorithm contains three sub-algorithms, namely pre-trained MAE, ConvNeXt, and SegFormer.  Based on the initials of these three sub-algorithms, the algorithm can be named MCS-DRNet.  Finally, we use the MCS-DRNet algorithm as an inspector to check and revise the results of the preliminary evaluation of the DR grade evaluation algorithm.  The experimental results show that the mean dice similarity coefficient of MCS-DRNet v1 and v2 are 0.5161 and 0.5544, respectively. The quadratic weighted kappa of the DR grading evaluation is 0.7559. Our code will be released soon.

\end{abstract}
\section{Introduction}
Diabetic retinopathy (DR) grade assessment is a very challenging task. Because in this work, it is always difficult for us to capture some very subtle DR feature in the image. However, capturing these subtle features is extremely important for assessing the severity of diabetic retinopathy. The traditional assessment of DR grading mainly relies on fundus photography and FFA, especially for PDR (proliferatived diabetic
retinopathy) and NPDR (non-proliferatived diabetic retinopathy). The main drawback of these tests is that they can cause serious harm to vision health. In addition, the FA detection method is mainly used to detect the presence or absence of neovascularization. It is often difficult to detect early or small neovascularization lesions by simply relying on fundus photography. Second, the test is an invasive fundus imaging method, which cannot be used in patients with allergies, pregnancy, and poor liver and kidney function. Therefore, for most cases, DR detection mainly adopts the ultra-wide OCTA method. This method can noninvasively detect the changes of DR neovascularization, which is an important imaging method for the ophthalmic diagnosis of PDR. However, at present, due to there being no work on automated DR analysis using standard ultra-wide (swept-source) optical coherence tomography angiography (UW-OCTA) \cite{bin_sheng_2022_6362349,dai2021deep,liu2022deepdrid,sheng2022overview}, the available UW-OCTA datasets are scarce and training data with label information is limited.

In recent years, to solve the problem of lack of labeled data, researchers have proposed many methods to alleviate the need for labeled data \cite{Wu2018UnsupervisedFL,Caron2019UnsupervisedPO,Doersch2015UnsupervisedVR,Srivastava2015UnsupervisedLO,chen2021exploring}. Among these methods, self-supervised learning is one of the most representative methods \cite{chen2021exploring,Wang2017TransitiveIF,khan2022contrastive,bao2021beit,caron2021emerging}. Self-supervised learning is mainly through pre-training on a large number of unlabeled data, and then mining the supervised features hidden in the unlabeled data. For example, the masked image modeling algorithm \cite{bao2021beit,MAE} in the domain of self-supervised learning. The algorithm randomly masks a part of the image in different ways. And then let the algorithm learn on its own how to recover the broken area. This method can make the algorithm learn useful algorithm weights for downstream tasks (classification \cite{madzin2014analysis,10.1504/ijcse.2020.111426}, semantic segmentation \cite{madzin2009feature,UNET_DEP}, etc.)  during the pre-training period, so as to improve the recognition accuracy of the algorithm in downstream tasks.

Nowadays, self-supervised learning has been widely used in various fields and has achieved good results \cite{zhai2019s4l}. However, the fly in the ointment is that in the process of self-supervised learning, it is difficult for an algorithm to efficiently dig out the supervised information beneficial to downstream tasks from a completely unfamiliar field without relying on any hints and only relying on its own intuition. For example, when we need to organize some unfamiliar and disorganized documents, we always find it very difficult. Because there are too many uncertainties here. However, if someone gives us information related to the classification of these documents. This may be of great help to us because it gives us a first glimpse of these unfamiliar files. 

To this end, in this paper, in order to mine supervised features contained in UW-OCTA images more efficiently, we construct a method between supervised learning and self-supervised learning, Briefly, this method is a suite of semi-supervised semantic segmentation methods for diabetic retinopathy grade assessment.  Overall, the method can be divided into the following two frameworks: 

The first framework is a semantic segmentation method of UW-OCTA diabetic retinopathy image with semi-supervised cross-algorithm ensemble. In this framework, first, we pre-train the MAE algorithm \cite{MAE} on the DR grade assessment dataset in the diabetic retinopathy analysis challenge (DRAC) 2022\footnote{https://drac22.grand-challenge.org/}. MAE algorithm \cite{MAE} as a common structure in self-supervised masked image modeling visual representation learning \cite{bao2021beit,MAE}. The algorithm can alleviate the labeling requirements of UW-OCTA images through self-supervised pre-training.  Although the MAE \cite{MAE} algorithm designed based on the vision transformer (ViT) \cite{VIT} architecture has good performance, the algorithm of this architecture outputs single-scale low-resolution features instead of multi-scale features. However, by fusing multi-scale UW-OCTA image features, our method can better cope with the problem of reduced recognition accuracy caused by the size change of the lesion area. To this end, we also deploy a no-position encoding and layered transformer encoder, SegFormer \cite{Segformer}. The SegFormer \cite{Segformer} algorithm is good at mining the features of the lesion area in the image at different scales. On the other hand, the lightweight All-MLP decoder design adopted by SegFormer \cite{Segformer} can generate powerful representations without the need for complex and computationally demanding modules. Finally, in order to form a good complement with MAE \cite{MAE} and SegFormer algorithm \cite{Segformer} based on the self-attention mechanism. We also integrate an algorithm for pure convolutional architecture based on the "sliding window" strategy (i.e., without self-attention), ConvNeXt\cite{convnext}. The introduction of the convolution architecture makes our method more comprehensively obtain the global features (long-distance dependencies) in UW-OCTA images. Therefore, in this framework, we construct a semi-supervised cross-algorithm ensemble method for lesion identification in UW-OCTA DR images based on MAE, ConvNeXt, and SegFormer algorithms\cite{MAE,Segformer,convnext,mmseg2020,mmselfsup2021}. Based on the initials of these three sub-algorithms, our method can be named \textit{MCS-DRNet}. There are two versions of our method in total, MCS-DRNet v1 and v2, corresponding to the challenge version and the post-challenge version, respectively.

The second framework is a grading assessment method for diabetic retinopathy based on a threshold inspection mechanism.  In this framework, we use the MCS-DRNet as an inspector to check and revise the results of the preliminary evaluation of the DR grade evaluation algorithm (EfficientNet v2 \cite{tan2021efficientnetv2}). To be specific, the inspector is mainly composed of the DR region segmentation pre-trained weights of MCS-DRNet and grading evaluation algorithm. In addition, we set up a set of threshold systems and evaluation rules for evaluating the severity of DR.

Finally, on the test set of the DRAC2022, our method exhibits strong performance.  In terms of DR semantic segmentation, the mean dice similarity coefficient of MCS-DRNet v1 and MCS-DRNet v2 are 0.5161 and 0.5544, respectively.  In terms of DR grading evaluation, the quadratic weighted kappa of the DR grading evaluation method based on the inspection mechanism is 0.7559, which brings a 2.99\% improvement compared to the baseline method (EfficientNet v2 \cite{tan2021efficientnetv2}).

\section{Approach}

\subsection{MCS-DRNet Series Methods}

\begin{figure}
\includegraphics[width=330pt,trim=50 0 70 80] {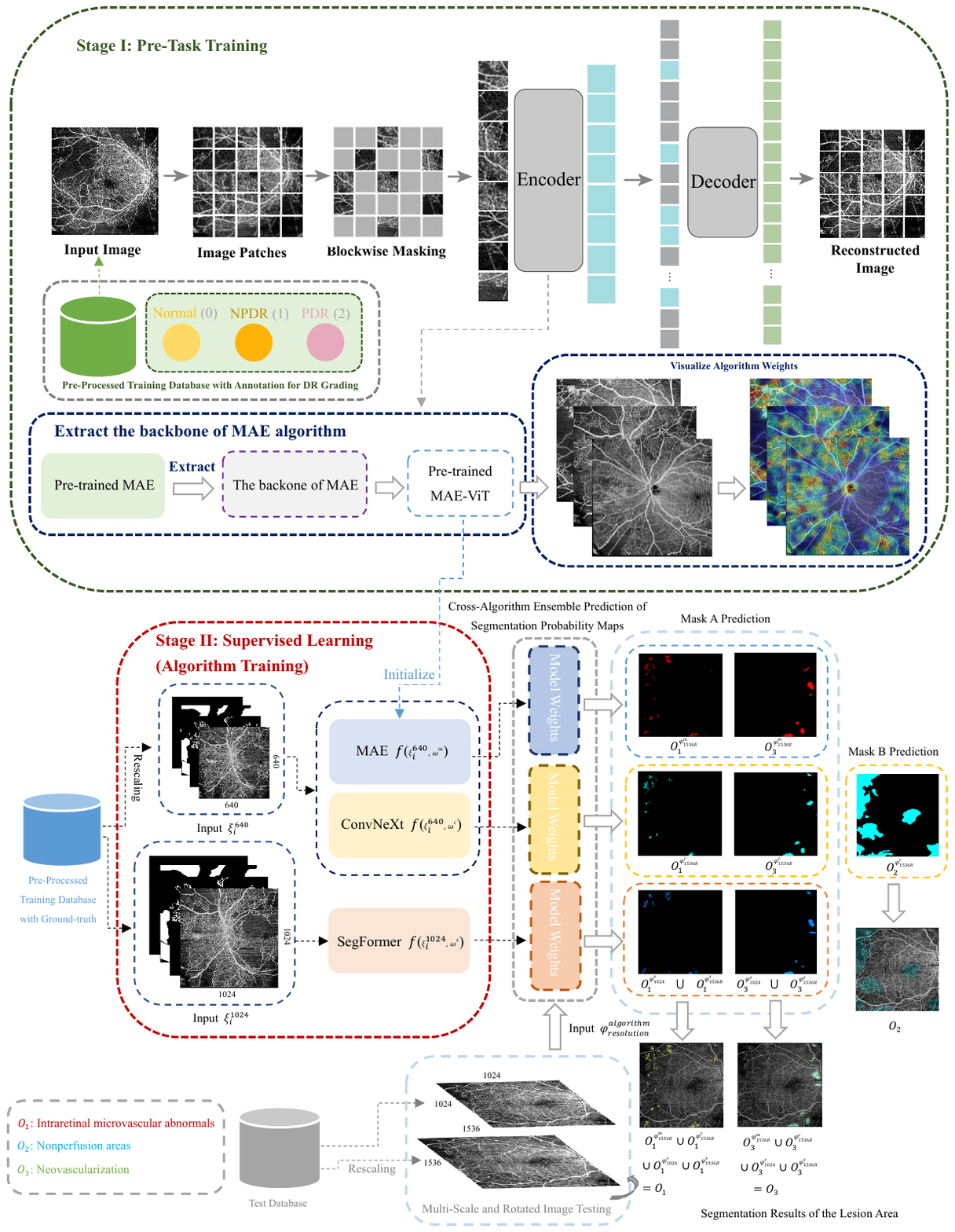}
\caption{The realization principle of MCS-DRNet method. Mask A indicates that intraretinal microvascular abnormalities and neovascularization are fused on a single mask for training, and Mask B indicates that the nonperfusion areas category is trained as a single mask.} \label{fig:1}
\end{figure}

\begin{figure}
\includegraphics[width=\textwidth,trim=54 500 50 225] {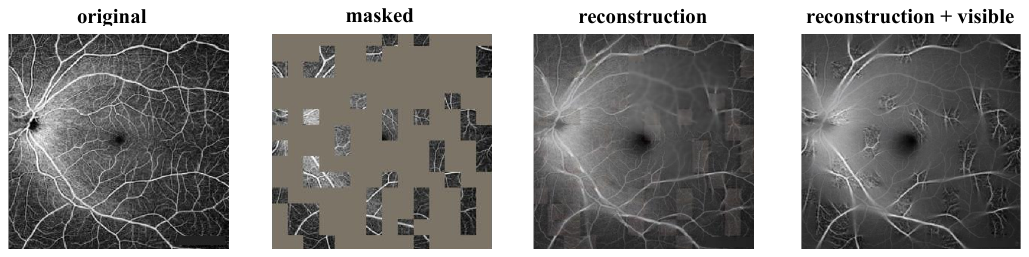}
\caption{The flow chart of the effect of MAE algorithm masking and reconstructing 75\% of the area in the UW-OCTA image.} \label{fig:2}
\end{figure}

The series method of MCS-DRNet we built can be divided into two stages, as shown in Figure \ref{fig:1}. These two phases are responsible for accomplishing two different tasks. The first stage is the pre-task training. This stage mainly relies on the MAE algorithm \cite{MAE} to pre-train on the diabetic retinopathy grade classification data set (Task 3) in the DRAC2022 challenge. In this process, for the MAE pre-training algorithm to comprehensively mine and learn the supervision information contained in the UW-OCTA image, the algorithm performs a masking operation on most of the areas in the input UW-OCTA image. The specific masking effect is shown in Figure \ref{fig:2}.


Then, the MAE encoder \cite{MAE} performs feature extraction on the visible patches subset of the image. In the process, the encoder will become a series of mask tokens. The full encoded patches and mask tokens are then processed by a small decoder. The main function of this small decoder is to reconstruct the original image in pixels. After the pre-training, we would extract the backbone of the MAE encoder  (also considered as pre-trained MAE-ViT) \cite{VIT,MAE}. And then, we use this extracted MAE-ViT as the initialization weight for the semantic segmentation task of the MAE algorithm. However, the MAE decoder is discarded \cite{MAE}.

The second stage is semantic segmentation learning (supervised learning) of UW-OCTA images. In this stage, in order to more fully explore the lesion features regions in UW-OCTA images, three algorithms with different visual processing strategies are deployed in the proposed method, which are pre-traind MAE \cite{MAE}, ConvNeXt \cite{convnext}, and SegFormer \cite{Segformer}. In this stage of learning, in order to speed up the whole training process, we set the training image size of the MAE \cite{MAE} and ConvNeXt \cite{convnext} algorithm to 640 × 640.

SegFormer \cite{Segformer}, a hierarchical transformer without position coding. Because the algorithm does not adopt the interpolation structure of position coding in the ViT algorithm, the algorithm can well cope with the influence brought by the recognition process of lesions with different sizes. Moreover, the high-resolution fine features and low-resolution coarse features generated by the encoder in SegFormer architecture can make up for the shortcoming that the MAE algorithm can only generate a single fixed-resolution low-resolution feature graphics to some extent. The MLP decoder part of the SegFormer algorithm \cite{Segformer} aggregates information from both lower and higher levels (lower levels tend to keep local features, while higher levels focus on global features). This enables the SegFormer network \cite{Segformer} to fuse multi-scale UW-OCTA image features in the process of processing UW-OCTA images, improve the perception field of the algorithm, and present a powerful representation.

ConvNeXt \cite{convnext}, a pure convolutional architecture algorithm (without Self-attention architecture). To some extent, this algorithm retains the "sliding window" strategy, which is beneficial to play the unique advantages of convolutional architecture in obtaining global image features (long-distance dependencies). This structural advantage can form a good complement with MAE \cite{MAE} and SegFormer \cite{Segformer} algorithms based on the self-attention mechanism.

After Stage I and Stage II, we can get the weights of MAE, ConvNeXt, and SegFormer algorithms. In the testing phase, we use a cross-algorithm ensemble to predict intraretinal microvascular abnormalities, nonperfusion areas, and neovascularization, the three categories of DR, and the predicted results corresponded to \(O_1\), \(O_2 \) and \(O_3 \) , respectively. Among them, part of the output adopts multi-scale (MS) prediction. In order to be able to obtain the union of image outputs of different sizes and output the final result, all the above 1536 × 1536 output sizes \{\(O_{\nu}^{\varphi_{1536 R}^{\lambda }}\),   \((\lambda  =m, c, s; \nu =1,3)\) and \((\lambda =c; \nu = 2)\) \}have been resized to 1024 × 1024 size (Note: the letter \(R \) after \(resolution\) is an abbreviation for \(rescaling\)).  Moreover, we found that a larger input size of the predicted image can improve the MAE and ConvNeXt algorithm's ability to capture subtle features. 

In conclusion, since there are many tiny features in the recognition of intraretinal microvascular abnormals and neovascularization, to better capture these features, both output items \(O_1\) and \(O_3\)  contain predicted values ensemble for multiple algorithms. Moreover, the output item \(O_{\nu}^{\varphi_{1536 R}^{i}}\)    contains the prediction results of the multi-angle (MA) rotated test image, and the realization principle is shown in Equation~\ref{eq:1}.

\begin{normalsize} 
\begin{equation} \label{eq:1}
O_{\nu}^{\varphi_{1536 R}^{\lambda }}=O_{\nu}^{\varphi_{1536 R}^{\lambda }} \cup f_{\tau  }\left(O_{\nu}^{\varphi_{1536 R}^{\lambda }}\right)
\end{equation} \end{normalsize}
Among them, the value of \(\nu\) is (\(1\),\(3\)). \(\tau \) stands for the angle of counterclockwise rotation, and its values are (\(90, 180, 270\)) degrees. There are two ways to assign \(i\), that is, when \(\lambda \) is (\(m, s\)), it corresponds to MCS-DRNet v1. When \(\lambda \) is (\(m, c, s\)), it corresponds to MCS-DRNet v2.

\begin{figure}
\includegraphics[width=\textwidth,trim=60 610 60 95] {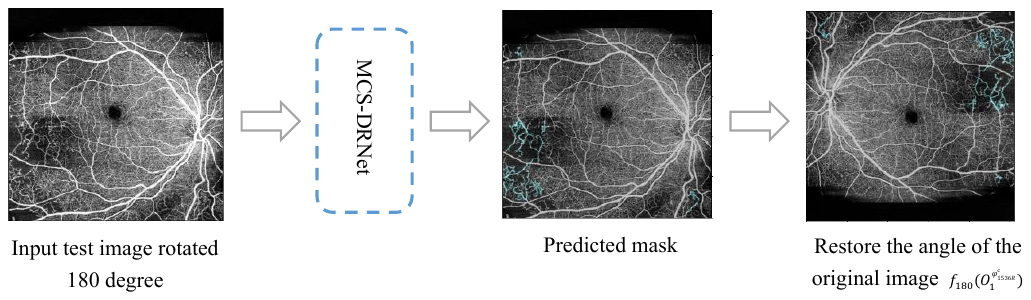}
\caption{The flow chart of the \(f_{180}\left(O_{1}^{\varphi_{1536 R}^{c}}\right)\)algorithm to rotate the image.} \label{fig:3}
\end{figure}

In order to show the implementation principle of the formula  \(f_{\tau}\left(O_{\nu}^{\varphi_{1536 R}^{\lambda }}\right)\) more concretely, we visualize how an example of this formula is implemented (in this example, \(\tau  = 180, \nu = 1, \lambda  = c.\)), as shown in Figure \ref{fig:3}.

For the output \(O_2\) of the nonperfusion areas, only \(O_{2}^{\varphi_{1536 R}^{c}}\) is included.

\begin{figure}[t]
\includegraphics[width=300pt,trim=10 20 90 10] {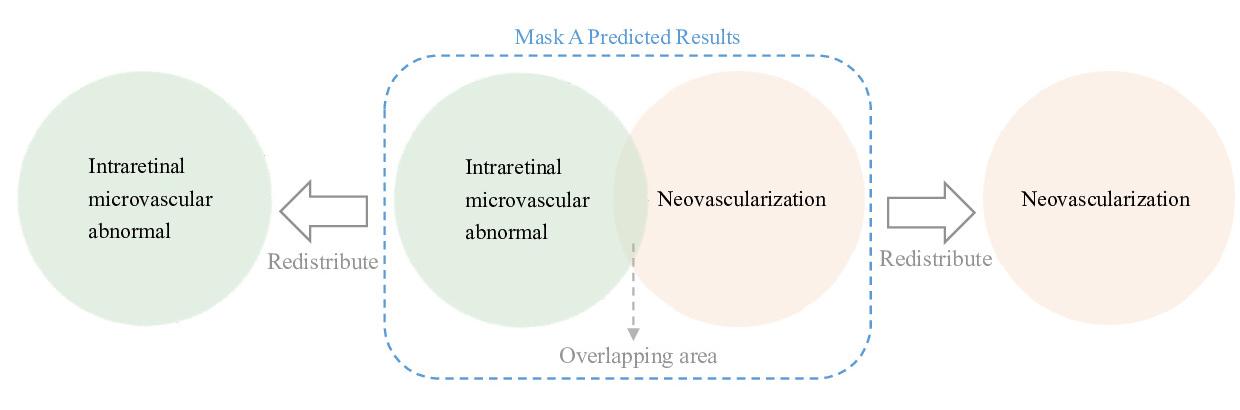}
\caption{The processing method of the overlapping area of Intraretinal microvascular abnormal and Neovascularization in the prediction result of Mask A} \label{fig:4}
\end{figure}

According to the above method, the two versions of the MCS-DRNet method constructed in this paper are as follows:

\subsubsection {MCS-DRNet v1 (Challenge Version)} In this version of the method, the version used by the ConvNeXt algorithm is L, and the algorithm is only used to predict Mask B. SegFormer and MAE algorithms are used for the prediction of Mask A. Therefore, the output items of MCS-DRNet v1 is shown in Equation ~\ref{eq:2}.

\begin{normalsize} 
\begin{equation} \label{eq:2}
\left\{\begin{array}{l}
O_{1}=O_{1}^{\varphi_{1536 R}^{m}} \cup O_{1}^{\varphi_{1024}^{s}} \cup O_{1}^{\varphi_{1536 R}^{s}} \\
O_{2}=O_{2}^{\varphi_{1536 R}^{c^{*}}} \\
O_{3}=O_{3}^{\varphi_{1536 R}^{m}} \cup O_{3}^{\varphi_{1024}^{s}} \cup O_{3}^{\varphi_{1536 R}^{s}}
\end{array}\right.
\end{equation} \end{normalsize}
The asterisk (\(*\)) in the upper right corner of the letter \(c\) indicates that the version of the algorithm ConvNeXt is L

\subsubsection {MCS-DRNet v2 (Post-challenge Version)} The method of this version is mainly based on the method of the previous version, with a slight improvement. The direction of improvement is reflected in: the version of the ConvNeXt algorithm is upgraded from L to XL. Furthermore, the algorithm is not only used for the prediction of Mask B, but also for the prediction of Mask A. Therefore, the output items of MCS-DRNet v2 is shown in Equation~\ref{eq:3}.

\begin{normalsize} 
\begin{equation} \label{eq:3}
\left\{\begin{array}{l}
O_{1}=O_{1}^{\varphi_{1536 R}^{m}} \cup O_{1}^{\varphi_{1536 R}^{c}} \cup O_{1}^{\varphi_{1024}^{s}} \cup O_{1}^{\varphi_{1536 R}^{s}} \\
O_{2}=O_{2}^{\varphi_{1536 R}^{c}} \\
O_{3}=O_{3}^{\varphi_{1536 R}^{m}} \cup O_{3}^{\varphi_{1536 R}^{c}} \cup O_{3}^{\varphi_{1024}^{s}} \cup O_{3}^{\varphi_{1536 R}^{s}}
\end{array}\right.
\end{equation} \end{normalsize}

\subsubsection {Post-processing} For the overlapping area of Intraretinal microvascular abnormal and Neovascularization in the prediction result of Mask A, our processing method is to directly distribute the overlapping area into the output results of Intraretinal microvascular abnormal and Neovascularization, respectively. The implementation process is shown in Figure \ref{fig:4}.

\subsection{DR Grade Assessment Method Based on Threshold Inspection Mechanism}

In image classification tasks, the features learned by a classification algorithm are always broad and holistic. However, the use of these broad features directly in the evaluation of the grade of retinopathy is inefficient. This is because some of the key diagnostic features of retinopathy are always subtle. Different from the classification task, in the learning of the semantic segmentation task, the algorithm can learn the ground-truth annotation in the DR image, so as to realize the acquisition of the detailed lesion features in the image. The limitation of semantic segmentation algorithms is that they cannot automatically classify images based on the learned features. For this reason, it seems to us that there is some kind of complementary relationship between them for the classification and segmentation tasks of UW-OCTA images. This relationship can be understood as subtle features learned in segmentation tasks can be used to examine and correct misclassified classes in classification tasks. Motivated by this idea, we constructed a method for assessing the grade of diabetic retinopathy based on a threshold inspection mechanism. The overall process of the method is shown in Figure \ref{fig:5}. As a whole, the method can be roughly divided into the following four steps:

\begin{figure}[tbh]
\includegraphics[width=250pt,trim=0 550 190 65] {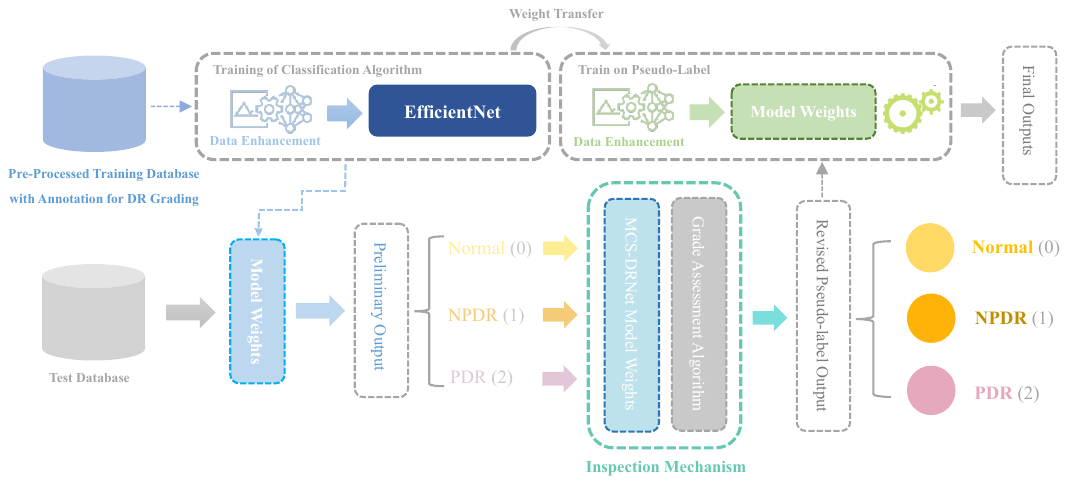}
\caption{Evaluation method of diabetic retinopathy grade based on threshold inspection mechanism} \label{fig:5}
\end{figure}

In the \textit{first} step (classification algorithm training), three data augmentation methods (MixUp \cite{zhang2017mixup}, CutMix \cite{yun2019cutmix}, and Crop) are used to avoid the EfficientNet algorithm \cite{tan2021efficientnetv2} from falling into the problem of overfitting.

In the \textit{second} step, we obtain the pretrained weights of the EfficientNet algorithm \cite{tan2021efficientnetv2} and perform DR grade assessment on the test images. After this step is completed, we can get the preliminary results of DR grade classification.

In the \textit{third} step, according to the weights of MCS-DRNet, we construct a grade revision mechanism based on pixel thresholds in the lesion area. The role of this mechanism is to adjust the results of the preliminary classification in the second step. In this mechanism, the MCS-DRNet method is responsible for identifying various lesion areas in the image, and accumulating and outputting the pixel values of these areas. One thing needs to be explained in particular, the MCS-DRNet method deployed in this framework is different from the MCS-DRNet v1 method. The main difference is that some of the output items are different. Equation \ref{eq:4} lists the output terms of the MCS-DRNet method adopted in this framework.

\begin{normalsize} 
\begin{equation} \label{eq:4}
\left\{\begin{array}{l}
O_{1}=O_{1}^{\varphi_{1536 R}^{m}} \cup O_{1}^{\varphi_{1024}^{s}} \cup O_{1}^{\varphi_{1536 R}^{s}} \\
O_{2}=O_{2}^{\varphi_{1536 R}^{m}} \\
O_{3}=O_{3}^{\varphi_{1536 R}^{m}} \cup O_{3}^{\varphi_{1024}^{s}} \cup O_{3}^{\varphi_{1536 R}^{s}}
\end{array}\right.
\end{equation} \end{normalsize}

\begin{table}
\caption{\(\mathcal{T}_{i}\) (and \( \mathcal{T}_{i}^{*}\)) is used to judge the pixel threshold of output \(O_i\) , where \(i\) equals (\(1, 2, 3\))}\label{tab1}
\centering
\begin{tabular}{|c|c|c|c|c|c|c|} 
\hline

Threshold & \(\mathcal{T}_{1}\)   & \(\mathcal{T}_{2}  \)                   & \(\mathcal{T}_{3}   \)                  & \(\mathcal{T}_{1}^{*}\)                     & \(\mathcal{T}_{2}^{*} \)                    & \(\mathcal{T}_{3}^{*} \)                      \\ 
\hline
Value     & 26\textsuperscript{2} & 130\textsuperscript{2} & 28\textsuperscript{2} & 78\textsuperscript{2} & 750\textsuperscript{2} & 100\textsuperscript{2}  \\

\hline
\end{tabular}
\end{table}

Based on Equation~\ref{eq:4}, we construct the inspection and revision mechanism as follows:

First, Equation \ref{eq:5} shows the sentence used to judge the pixel threshold condition judgment of each output item (\(O_1, O_2, O_3\)).

\begin{normalsize} 
\begin{equation} \label{eq:5}
\left\{\begin{array}{l}
C_{i}^{\min }: O_{i}<\mathcal{T}_{i} \\
C_{i}^{\max }: O_{i}>\mathcal{T}_{i}^{*}   
\end{array}\right.
\end{equation} \end{normalsize}

The conditional statements \({C}_i^{min} \) and \({C}_i^{max}\) are mainly used for the diagnosis of disease severity, which correspond to Normal and PDR respectively. In Equation \ref{eq:5}, the values of \(i\) are (\(1, 2, 3\)), Table \ref{tab1} gives the thresholds used for pixel judgment in each image.

\begin{figure}[tbh]
\includegraphics[width=240pt,trim=0 350 170 35] {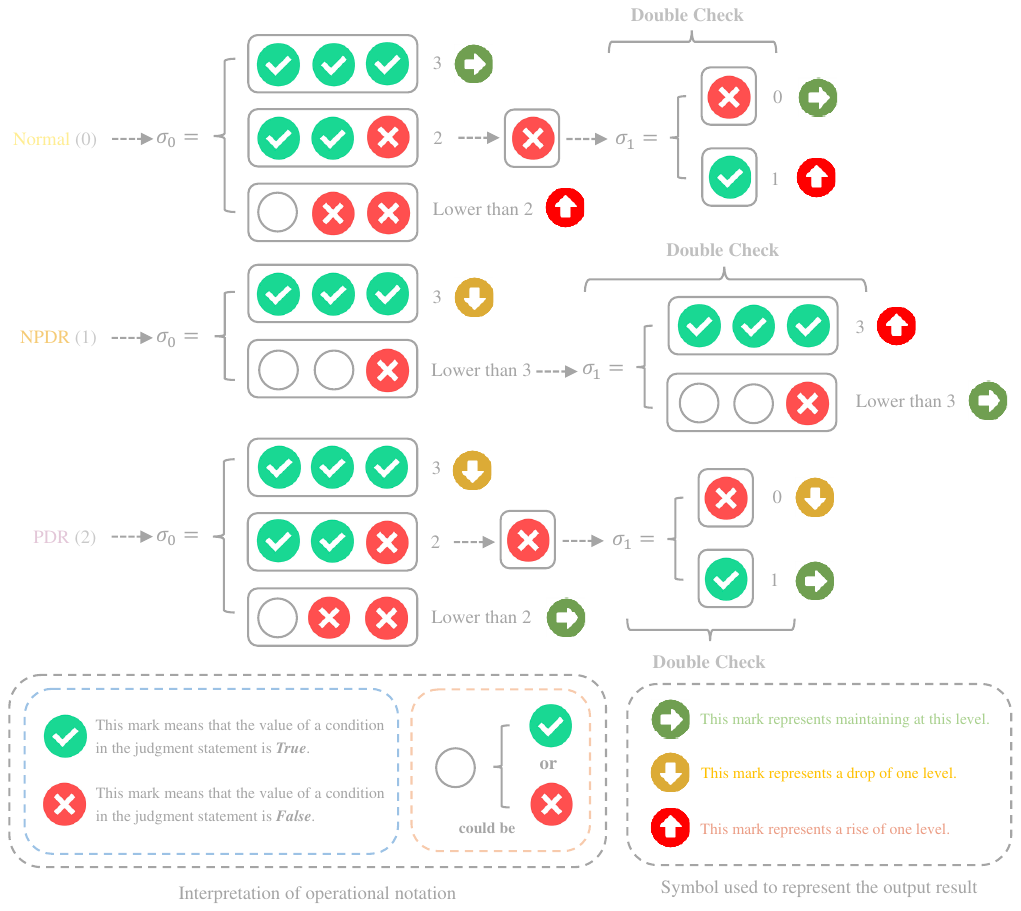}
\caption{The specific principle of the threshold inspection mechanism. The checks and adjustments for Normal (0), NPDR (1) and PDR (2) are in parallel, that is, the results of the adjustments do not affect each other. There is no ordering relationship between the elements (correct and incorrect symbols) within each parenthesis. All adjustments will only be made between adjacent levels.} \label{fig:6}
\end{figure}

Secondly, Equation \ref{eq:6} shows the cumulative calculation formula of the judgment conditions true and false.

\begin{normalsize} 
\begin{equation} \label{eq:6}
f_{c}(x)=\left\{\begin{array}{l}
if \ x \ equals \ True, return \ 1   \\
 if \ x \ equals \ False, \ return \ 0
\end{array}\right.
\end{equation} \end{normalsize}

Finally, Equations \ref{eq:7} and \ref{eq:8} give the overall calculation formula of the pixel threshold inspection mechanism.

\begin{normalsize} 
\begin{equation} \label{eq:7}
\sigma_{0}=\sum_{i=1}^{3} f_{\mathrm{c}}\left(C_{i}^{\min }\right)
\end{equation} \end{normalsize}

\begin{normalsize} 
\begin{equation} \label{eq:8}
\sigma_{1}=\sum_{i=1}^{3} f_{\mathrm{c}}\left(C_{i}^{\max }\right)
\end{equation}
\end{normalsize}

Based on Equations \ref{eq:7} and \ref{eq:8}, we constructed the diabetic retinopathy grade inspection and adjustment mechanism, as shown in Figure \ref{fig:6}.

For the Normal (0) level, our adjustment scheme is as follows:

(1) \(\sigma_0=\ 3\), which means that the sample satisfies all the conditions for the level of Normal, so it remains unchanged at this level;

(2) \(\sigma_0=\ 2\), which means that the sample only satisfies some of the conditions for the Normal level. We need further confirmation of that \(False\) condition. This False condition can be identified as NPDR as long as it satisfies any of the conditions in the \( \sigma_1\) judgment statement, instead, it remains unchanged at that level.

(3) \(\sigma_0\le\ 1\), which means that the sample is not enough to be recognized as Normal. Therefore, the sample needs to be upgraded to NPDR.

For the NPDR (1) level, our adjustment scheme is as follows:

(1) \(\sigma_0=\ 3\) , which means that the sample satisfies all conditions for level Normal. Therefore, this sample is downgraded to Normal.

(2) \(\sigma_0\le\ 2\), which means that it is reasonable for this sample to be considered NPDR , but further confirmation is required. If the sample meets all the conditions of \( {\sigma}_1\), it is deemed unreasonable at the NPDR  level and will be upgraded to PDR, otherwise, it will remain at that level.

For the PDR (2) level, our adjustment scheme is as follows:

(1) \(\sigma_0=\ 3\), which means that the sample does not meet the level of PDR and needs to be lowered by one level.

(2) \(\sigma_0=\ 2\), which means that the sample may not meet the level of PDR, We need further confirmation of that \(False\) condition. As long as this \(False\) condition, in the judgment statement \( {\sigma}_1\), satisfies a condition, it can be considered that the sample is divided into PDR is reasonable, otherwise, it is unreasonable

(3) \(\sigma_0\le\ 1\), which means that the sample is reasonable at the level of PDR, so it will be retained at this level.

In the \textit{last} step (step 4), the adjusted pseudo-annotations are used to train EfficientNet v2, and the final result is output. In this step, the methods of data enhancement are MixUp \cite{zhang2017mixup} , and CutMix \cite{yun2019cutmix}.

\section{Experimental Settings and Results}

\subsection{Segmentation of Diabetic Retinopathy Tissue}
In the experimental framework for segmentation of diabetic retinopathy tissue, we first do semi-supervised pre-training in the DRAC2022 DR grading assessment training set. Then, on the training set of DRAC2022 semantic segmentation, we perform supervised training to evaluate the representation by end-to-end fine-tuning. In addition, the results obtained by various experimental methods on the segmentation leaderboard are presented. Finally, to further verify the effectiveness of the proposed method, we visualize the semantic segmentation results of the three DR categories (intraretinal microvascular abnormals, nonperfusion areas, and neovascularization) on the test set.

\begin{table}[t]
\caption{Data augmentation results of DRAC2022 DR grading dataset. 0 to 2 represent the three DR levels of Normal, NPDR, and PDR in turn.}\label{tab2}
\centering
\begin{tabular}{|c|c|c|c|c|} 
\hline
Level of assessment                            & 0    & 1    & 2   & Total  \\ 
\hline
Amount
  of raw data               & 328  & 213  & 70  & 611    \\ 

Quantity
  after data augmentation & 1968 & 1278 & 420 & 3666   \\
\hline
\end{tabular}
\end{table}

\begin{table}[h]
\caption{Data augmentation results of the DRAC2022 semantic segmentation dataset. The data is enhanced by flipping vertically and horizontally and rotating counterclockwise 90 degrees, 180 degrees, 270 degrees. Since nonperfusion areas in the image has more pixel values than classes intraretinal microvascular abnormalities and neovascularization. In order to avoid the class unbalance problem that may be caused by the algorithm in the identification process of each category, we write the ground truth of nonperfusion areas to Mask B, and the ground truth of the remaining two lesion types is written on Mask A.}\label{tab3}
\centering
\begin{tabular}{|c|c|cc|} 
\hline
\multicolumn{2}{|c|}{Types of mask }                    & \multicolumn{0}{c}{A} & B    \\ 
\hline

\multirow{2}{*}{The original data in the data set~ } & Number of images with lesions present  & 88                     & 106  \\ 

                                                     & Number of images without lesions   & 21                     & 3    \\ 
\hline
\multirow{2}{*}{Quantity after data augmentation }   & Number of images with lesions present & 528                    & 636  \\ 

                                                     & Number of images without lesions   & 126                    & 18   \\
\hline
\end{tabular}
\end{table}

\subsubsection {Semi-Supervised Pre-training  (Classification Algorithm Training)}
In the semi-supervised pre-training part of MAE \cite{MAE}, to make the pre-training algorithm more fully mine the supervised information contained in UW-OCTA images, so as to improve the performance of the deep learning algorithm in processing semantic segmentation task. We use the DR assessment section of the DRAC2022 dataset. The  dataset contains a total of 611 original images. We process these images by rotating 90 degrees, 180 degrees, and 270 degrees as well as flipping horizontally and vertically to acquire a total of 3666 images for pre-training, as shown in Table \ref{tab2}. In addition, we do various data enhancements during pre-training, including brightness, contrast, saturation and hue. 

The training parameters of MAE algorithms are as follows: backbone (vit-base); resized crop (480); base learning rate (1.5e-4); warm-up iteration (40); max epochs (1600).

\subsubsection{Supervised Learning (Segmentation Algorithm Training)}

The original data of the semantic segmentation dataset provided by the DRAC2022 challenge has 109 images with a pixel value of 1024 x 1024. It includes the following three diagnostic categories: intraretinal microvascular abnormals, nonperfusion areas, and neovascularization. In the image data of these three categories, there are  86, 106, and 35 images with manual annotation (lesions in the images), respectively. In the process of the experiment, to avoid the algorithm falling into over-fitting and improve the generalization of the algorithm, we augment 109 images in all the original data, and the augmentation results are shown in Table \ref{tab3}. The image data of each category are angle-adjusted and expanded to a total of 654 images (including the original image). We believe that in the training process, adding images of non-diseased areas (background images) is conducive to enhancing the learning of the deep learning algorithm for non-diseased areas and reducing the probability of misjudgment of the algorithm for diseased areas. In addition, we also do various data enhancements during the training, including brightness, contrast, saturation and hue.

In algorithm training , we train three algorithms, which are pre-trained MAE, SegFormer, and ConvNeXt. The training parameters of these algorithms are as follows:

(1) The parameters of pre-trained MAE algorithm training are as follows: pre-training algorithm (vit-base); crop size (\(640^{2}\)); base learning rate (1e-6); optimizer (AdamW); max iterations (130k); loss (CrossEntropyLoss).

(2) The parameters of SegFormer algorithm training are as follows: pre-training algorithm (mit-b3); crop size (\(1024^{2}\)); base learning rate (1e-6); optimizer (AdamW); max iterations (130k); loss (CrossEntropyLoss).

(3) The parameters of ConvNeXt algorithm training are as follows: pre-training algorithm (ConvNeXt L/XL); crop size (\(640^{2}\)); base learning rate (8e-6); optimizer (AdamW); max iterations (110k for L / 130k for XL); loss (CrossEntropyLoss).

\subsubsection{Experimental Results}
For quantitative evaluation, Table \ref{tab4} shows the mean Intersection over Union (mIoU) and mean Dice (mDice) of subclass 1 and 3 for algorithms UNet + DeepLabV3 \cite{UNET_DEP} and MCS-DRNet series methods in the DRAC2022 semantic segmentation data set. As you can see from the table, the MCS-DRNet series of methods constructed in this paper significantly outperformed the traditional medical semantic segmentation algorithm UNet + DeepLabV3 \cite{UNET_DEP} in identifying the mIoU and mDice of two subcategories (intraretinal microvascular abnormals and neovascularization) .

Among the MCS-DRNet family of algorithms, the original MCS-DRNet v2 (\(A.3\)) method achieves 5.26\% and 0.39\% improvement on the mDice metric for categories 1 and 3, respectively, compared to v1.  Moreover, in the original MCS-DRNet v2 method, when we add multi-scale (MS) test image operation (\(A.4\)), the mDice measurement index of Classes 1 and 3 is improved by 0.1\% and 0.44\% respectively.  This means that this operation can effectively improve the performance of our method in the identification of DR areas at multiple scales. Based on performing the MS, we added the multi-angle rotation (MR) operation to obtain our final MCS-DRNet v2 method (\(A.5\)).  The addition of this method increased the mDice index of Classes 1 and 3 by 2.87\% and 0.16\%, respectively.

Table \ref{tab5} presents the mIoU and mDice of UNet+DeepLabV3 and MCS-DRNet series methods in identifying Class 2 in the DRAC2022 semantic segmentation dataset. As can be seen from the table, MCS-DRNet v2 based on ConvNeXt-XL (\(B.4\)) achieves the best results in both mIoU and mDice metrics in the task of identifying Class 2. However, when we conduct MR operations on the MCS-DRNet v2 method (\(B.3\)), the measured values of mIoU and mDice both decreased to varying degrees.

\begin{table}[h]
\caption{Comparison of the performance of MCS-DRNet in identifying intraretinal microvascular abnormalities (Class 1) and neovascularization (Class 3) of lesion classes. In the parameter Image Size, M represents the existence of two resolutions (640×640 and 1024×1024) in this method. In the parameter Backbone, the letters M, C, and S indicate the integration of MAE (ViT-base) \cite{MAE}, ConvNeXt-XL\cite{convnext} and SegFormer-B3 \cite{Segformer}, respectively. \(w/o\) is short for \(without\).}\label{tab4}
\centering
\resizebox{\textwidth}{!}{
\begin{tabular}{|c|l|c|c|c|c|c|c|} 
\hline
    &                       &             &            & \multicolumn{4}{c|}{Mask A}                                        \\ 
\cline{5-8}
ID  & \multicolumn{1}{c|}{Method}                    & Backbone    & Image Size & \multicolumn{2}{c|}{Class 1}    & \multicolumn{2}{c|}{Class 3}     \\ 
\cline{5-8}
    &                       &             &            & mIoU(\%)       & mDice(\%)      & mIoU(\%)       & mDice(\%)       \\ 
\hline
A.1 & UNet +
  DeepLabV3 \cite{UNET_DEP}    & UNet-S5-D16 & 1024×1024  & 13.85          & 22.35          & 33.80          & 48.39           \\ 

A.2 & MCS-DRNet v1             & MS          & M          & 22.83          & 35.79          & 41.06          & 57.04           \\ 

A.3 & MCS-DRNet v2 (\(w/o\) MR \& MS)             & MCS         & M          & 26.98          & 41.05          & 41.56          & 57.43           \\ 

A.4 & MCS-DRNet v2 (\(w/o\) MR)   & MCS         & M          & 27.30          & 41.15          & 41.86          & 57.87           \\ 

A.5 & MCS-DRNet v2 & MCS         & M          & \textbf{29.63} & \textbf{44.02} & \textbf{42.13} & \textbf{58.03}  \\
\hline
\end{tabular}}
\end{table}

\begin{table}
\caption{Performance comparison of various algorithms in identifying nonperfusion areas (Class 2/ Mask B).  \(w\) is short for \(with\).}\label{tab5}
\centering
\begin{tabular}{|c|l|c|c|c|c|} 
\hline
\multicolumn{1}{|c|}{ID} & \multicolumn{1}{c|}{Method}             & Backbone    & Image
  Size & mIoU(\%)       & mDice(\%)       \\ 
\hline
B.1                      & UNet + DeepLabV3 \cite{UNET_DEP} & UNet-S5-D16 & 1024×1024    & 41.06          & 55.50           \\ 

B.2                      & MCS-DRNet v1          & ConvNeXt-L  & 640 × 640    & 47.78          & 61.98           \\ 

B.3                      & MCS-DRNet v2 (\(w \) MR)  & ConvNeXt-XL & 640 × 640    & 48.74 & 62.96\\ 

B.4                      & MCS-DRNet v2    & ConvNeXt-XL & 640 × 640    & \textbf{49.99} & \textbf{64.26}  \\
\hline
\end{tabular}
\end{table}

Finally, Table \ref{tab6} shows the final mean dice similarity coefficient (mean DSC) obtained by all methods. It can be seen that the MCS-DRNet v2 (composed of \(A.5\) and \(B.4\)) outperforms other methods.

\begin{table}
\caption{Comparison of Mean DSC for each model and algorithm. The \textit{asterisk} (*) in the upper right corner of the number represents the final result of the DRAC2022 challenge. The following table follows this principle.}\label{tab6}
\centering
\begin{tabular}{|c|c|c|c|} 
\hline
ID                        & Backbone
  (Mask A \& B)      & mean DSC
  (\%)  \\ 
\hline
A.1 \& B.1   & UNet-S5-D16                 & 42.08            \\ 

A.2 \& B.2        & MS \& ConvNeXt-L            & 51.61\^{*}             \\ 

A.3 \& B.4          & MCS \& ConvNeXt-XL          & 54.23            \\ 

A.4 \& B.4     & MCS  \& ConvNeXt-XL          & 54.41            \\ 

A.5 \& B.4  & MCS \& ConvNeXt-XL & \textbf{55.44}   \\
\hline
\end{tabular}
\end{table}

\begin{figure}[t]
\includegraphics[width=275pt,trim=55 340 150 95] {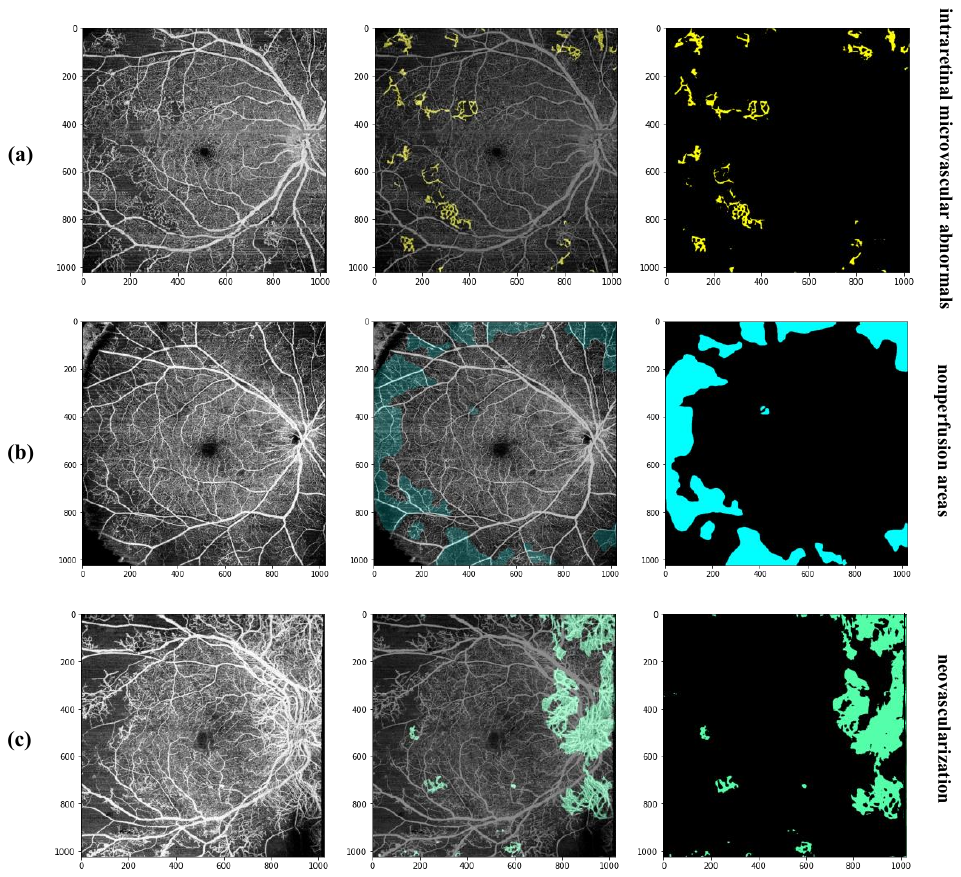}
\caption{The recognition effect of the method in this paper in three sub-categories of lesion areas.} \label{fig:7}
\end{figure}

In order to further verify the effectiveness of MCS-DRNet series methods in this paper. Figure \ref{fig:7} (a) - (c) sequentially shows the recognition effects of the proposed method on three DR lesion categories (intraretinal microvascular abnormalities, nonperfusion areas and neovascularization) on the test set of DRAC2022. From the images, it can be seen that MCS-DRNet can well identify the regions of various DR areas in UW-OCTA images.

\begin{figure}
\includegraphics[width=\textwidth,trim=90 195 100 100] {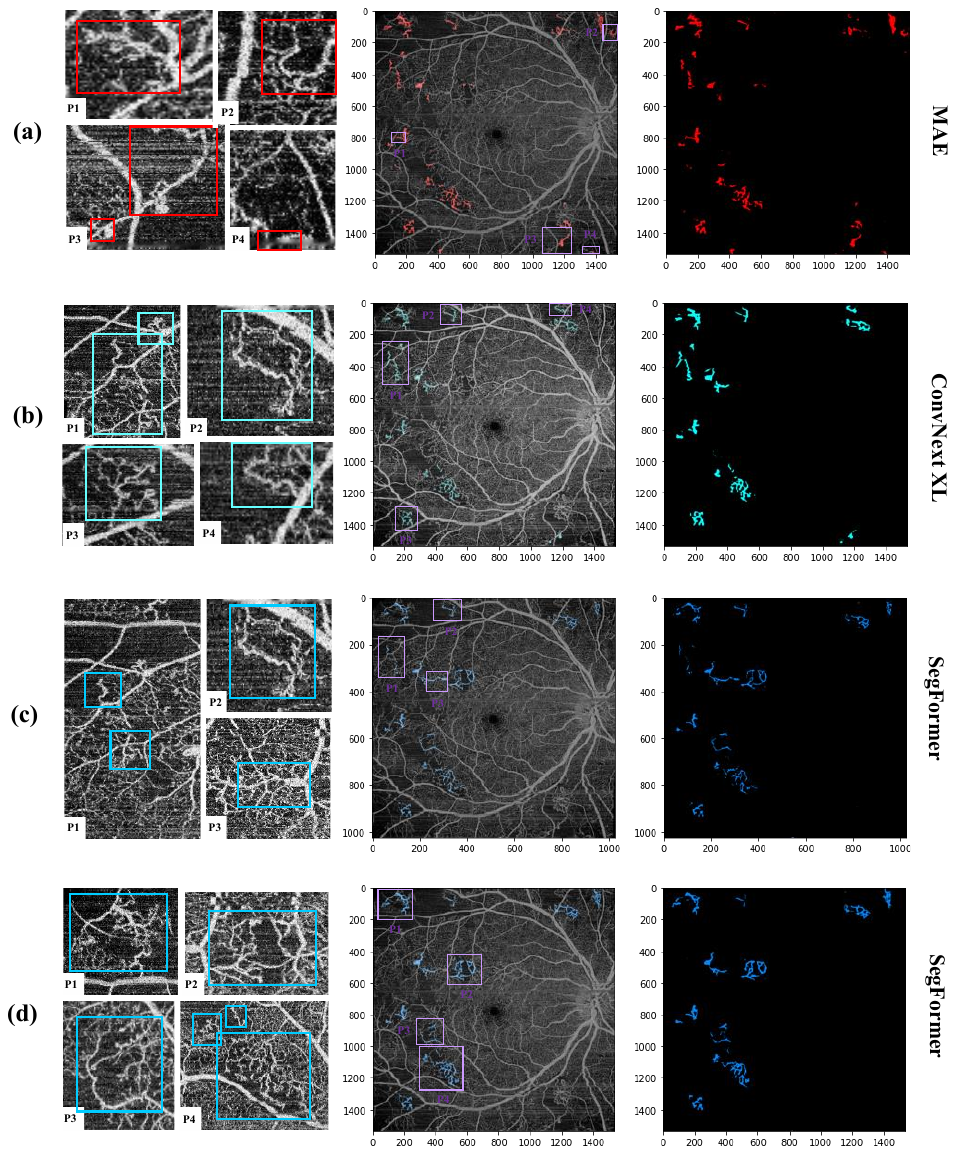}
\caption{The identification effect of each sub-algorithm in intraretinal microvascular abnormals lesion category. (a) and (b) are the test image recognition effect of MAE and ConvNeXt at a resolution of 1536×1536, respectively. (c) and (d) represent the recognition effect of SegFormer on the scale of 1024×1024 and 1536×1536, respectively. P1- P4 represent patches 1-4 in turn. P1 - P4 in the leftmost column corresponds to the enlargement of P1 - P4 in the middle column. The groundtruth with different colors in each Patch in the leftmost column represents the area where the diseased tissue is located. Figure \ref{fig:9} also follows these principles.} \label{fig:8}
\end{figure}

\begin{figure}
\includegraphics[width=\textwidth,trim=85 195 100 110] {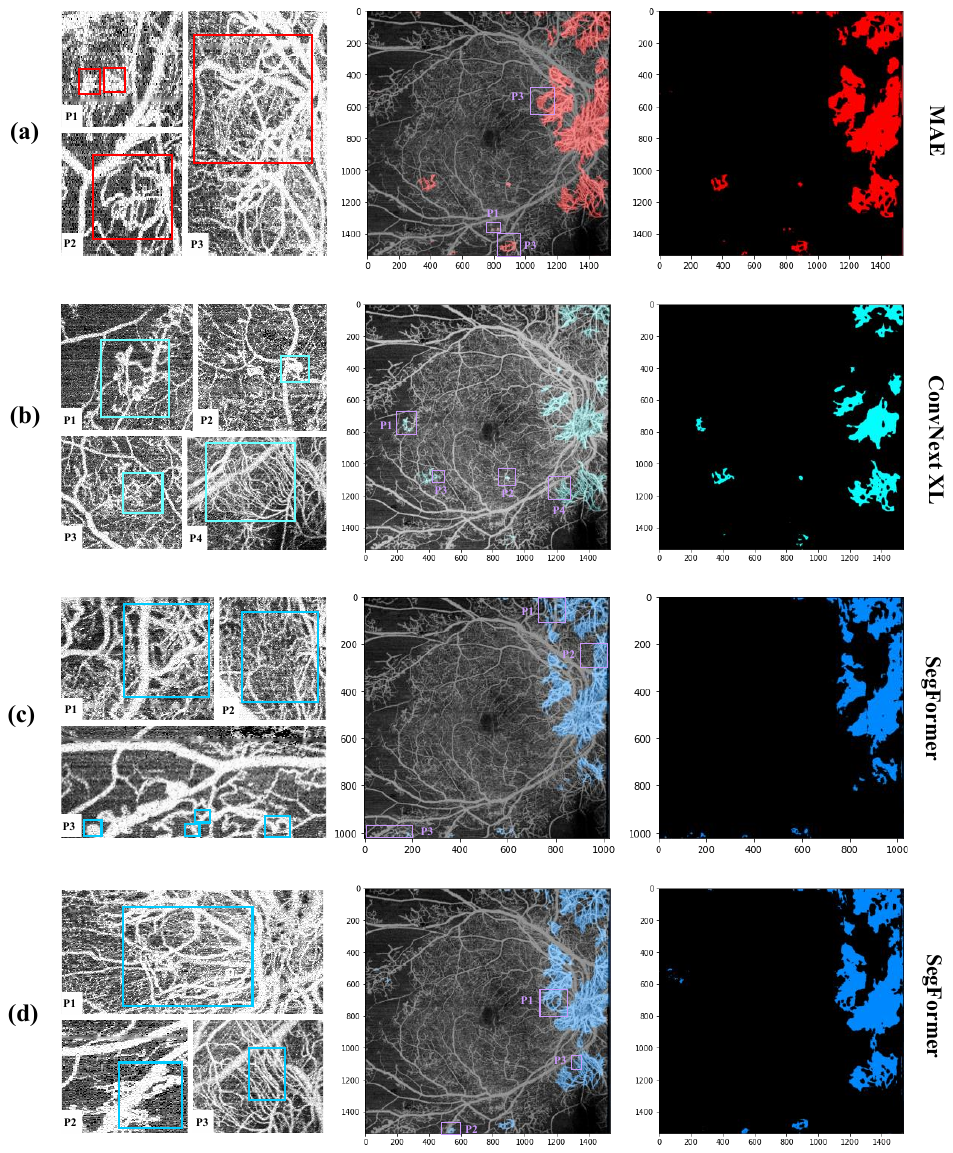}
\caption{Recognition effect of each sub-algorithm in neovascularization lesion category. This diagram follows the same principles as Figure \ref{fig:8}.} \label{fig:9}
\end{figure}

In order to more intuitively show the various algorithms in the proposed method, and the role play in these identification results. Figure \ref{fig:8} and Figure \ref{fig:9} demonstrate the recognition effect of these sub-algorithms on intraretinal microvascular abnormals and neovascularization for DR categories, respectively. In these two images, we annotate the unique contribution (patches 1-4) of each sub-algorithm in the recognition of pathological tissue features in the whole image.

From Figure \ref{fig:8}, it is not difficult to find that the MAE algorithm based on the self-attention structure can well excavate some unique subtle DR tissue. The ConvNeXt algorithm based on the "sliding window vision" strategy can find DR tissue in the image more evenly. The segFormer algorithm plays an important role in the recognition of lesions of different scales in images.

From Figure \ref{fig:9}, we can see that the MAE algorithm and SegFormer based on the self-attention architecture can pay attention to the most obvious areas in the image, and at the same time, they can also pay attention to some subtle lesion features. The ConvNeXt algorithm based on the sliding window strategy can still have a general grasp of the global features (long-distance dependencies). SegFormer is very robust in the recognition of neovascularization lesions of different sizes.

Therefore, the set of sub-algorithms with different visual strategies constructed in this paper have their own unique contributions in identifying DR tissue. This result further verifies the rationality of the hypothesis in this paper, that is, our method integrating different visual processing strategies can more efficiently mine the diseased tissue contained in the image.

\subsection {Diabetic Retinopathy Grade Assessment}

In the experimental framework of DR grade evaluation, first, we perform supervised learning (classification algorithm training) on the DRAC2022 DR grade evaluation data set to obtain the results of the preliminary evaluation. Then, we adjust the results of the preliminary evaluation through a threshold inspection mechanism to obtain pseudo-annotations. Finally, we use the pseudo-annotations to fine-tune the weights of the algorithm and get the final result.

\subsubsection {Classification Algorithm Training }

The training of the EfficientNet v2 \cite{tan2021efficientnetv2} classification algorithm is similar to the semi-supervised pre-training of the MAE \cite{MAE} algorithm, which is essentially a classification task and shares the same dataset. Therefore, in the training process of the classification algorithm, the expansion results of the training data are consistent with the results in Table \ref{tab2}. 

\begin{table}[t]
\caption{ The effect of TIM method for DR Grade evaluation. }\label{tab7}
\centering
\begin{tabular}{|c|c|} 
\hline
Method                    & Quadratic weighted kappa
  (\%)  \\ 
\hline
EfficientNet v2 \cite{tan2021efficientnetv2}       & 72.60                              \\ 

EfficientNet v2 + TIM  & \textbf{75.59}\^{*}                            \\
\hline
\end{tabular}
\end{table}

Overall, our algorithm training can be divided into two parts. 

(1) The first part is the training based on the DR Level assessment data set. The training parameters of this part are as follows: experimental algorithm (EfficientNet v2 M); initialization of weights (ImageNet); max epochs (220); learning rate warmup epoch (40); dropout (0.5); data enhancement (Mixup, CutMix, and Crop). 

(2) The second part is the training based on the pseudo-labeled dataset. The training parameters of this part are as follows: experimental algorithm (Efficientnet v2 M); initialization of weights (the weight of the first part of the algorithm); max epochs (100); learning rate warmup epoch (10); dropout (0.2); data enhancement (Mixup and Cutmix). 

The classification algorithm training image size of these two parts is \(480^{2}\).

\subsubsection {Experimental results and analysis}
For quantitative evaluation, Table \ref{tab7} lists the results obtained by EfficientNet v2 \cite{tan2021efficientnetv2} and EfficientNet v2 + threshold inspection mechanism (TIM) on the competition list. As can be seen from the table, after using the TIM method, the EfficientNet v2 brings a 2.99 \% improvement in quadratic weighted kappa. To some extent, the results verify the rationality of our conjecture about the difference between classification and segmentation algorithms in obtaining UW-OCTA image features. 

Finally, we summarize three shortcomings of the framework. First of all,  DR grading assessment method based on a threshold value, and the selection of threshold value is highly professional. If the threshold selection is good, the performance of the algorithm to identify the DR Organization could be greatly improved. Second, in this challenge, due to the urgency of time, the semantic segmentation model we built for grade evaluation is relatively simple and needs to be improved. Last but not least, in this challenge, the method we submitted in the system only performed one iteration between the \textit{third} step and the \textit{fourth} step. However, multiple iterations can be beneficial for more optimal results.

\section{Conclusion}

Typically, microvascular hemorrhage results in the formation of new blood vessels that subsequently transition to proliferative DR.  This is potentially harmful to vision.  To facilitate the development of diabetic retinopathy automatic detection, this paper proposes a novel semi-supervised semantic segmentation method for UW-OCTA DR grade assessment. In this method, firstly, supervised DR feature information in UW-OCTA images is mined through semi-supervised pre-training. Secondly, a cross-algorithm integrated DR Semantic segmentation algorithm is constructed according to these characteristics. Finally, a DR grading evaluation method is constructed based on this semantic segmentation algorithm. The experimental results verify that our method can accurately segment most of the images classified as proliferative DR and assess the DR Level reasonably.

\bibliographystyle{splncs04} 
\bibliography{cas-refs.bib}

\end{document}